\documentclass[conference]{IEEEtran}
\IEEEoverridecommandlockouts
\usepackage{cite}
\usepackage{amsmath,amssymb,amsfonts}
\usepackage{algorithmic}
\usepackage{graphicx}
\usepackage{textcomp}
\usepackage{xcolor}
\def\BibTeX{{\rm B\kern-.05em{\sc i\kern-.025em b}\kern-.08em
    T\kern-.1667em\lower.7ex\hbox{E}\kern-.125emX}}
\begin{document}

\title{Supervised learning for crop/weed classification based on color and texture features}

\author{\IEEEauthorblockN{ Faiza Mekhalfa}
\IEEEauthorblockA{\textit{Division Productique et Robotique} \\
\textit{Centre de Developpement de Technologies Avancees}\\
Algiers, Algeria \\
fmekhalfa@cdta.dz}
\and
\IEEEauthorblockN{ Fouad Yacef}
\IEEEauthorblockA{\textit{Division Productique et Robotique} \\
\textit{Centre de Developpement de Technologies Avancees}\\
Algiers, Algeria \\
fyacef@cdta.dz}

}

\maketitle

\begin{abstract}
Computer vision techniques have attracted a great interest in precision agriculture, recently. The common goal of all computer vision-based precision agriculture tasks is to detect the objects of interest (e.g., crop, weed) and discriminating them from the background. The Weeds are unwanted plants growing among crops competing for nutrients, water, and sunlight, causing losses to crop yields. Weed detection and mapping is critical for site-specific weed management to reduce the cost of labor and impact of herbicides. This paper investigates the use of color and texture features for discrimination of Soybean crops and weeds. Feature extraction methods including two color spaces (RGB, HSV), gray level Co-occurrence matrix (GLCM), and Local Binary Pattern (LBP) are used to train the Support Vector Machine (SVM) classifier. The experiment was carried out on image dataset of soybean crop, obtained from an unmanned aerial vehicle (UAV), which is publicly available.  The results from the experiment showed that the highest accuracy (above $96$\%) was obtained from the combination of color and LBP features.
\end{abstract}


\section{Introduction}
In recent years, there has been a strong activity in precision agriculture (PA), particularly the monitoring aspect. Precision agriculture employs data from multiple sources for the purpose of improving crop yields and increasing the cost-effectiveness of crop management strategies including fertilizer inputs, irrigation management, and pesticide application \cite{b1}. PA offers the opportunity for a farmer to apply the right amount of treatment at the right time and at the right place \cite{b2}. Nowadays, (UAVs) can be exploited in a variety of applications related to crops management, by capturing high spatial and temporal resolution images of the entire agricultural field. UAVs have been considered more efficient, compared to the ground robot or satellite acquisitions, since they allow a fast acquisition of the field with very high spatial resolution and at a low cost \cite{b3}.

Among the most popular application of UAVs in Precision Agriculture is weed mapping \cite{b4}\cite{b5}. Weeds are undesirable plants, which grow in agricultural crops and can cause several problems. They are competing for available resources such as water or even space, causing losses to crop yields \cite{b3}. The knowledge of weed infestation is an essential procedure for the use of preventive measures in their control. The challenge of crop/weed classification was addressed by considering various machine learning techniques. The most popular classification techniques are the Artificial Neural Networks (ANNs) family \cite{b6} and the Random Forest algorithm \cite{b7}.

Based on  the  principle  of  structure  minimum  according  to  the  statistical   learning   theory developed by Vapnik and Chervonenkis \cite{b8},  Support Vector Machines (SVMs)  can  solve  practical  problems  encountered  with  traditional  classifiers  in  the  aspects   of   small   training   samples,   nonlinearity,   high   dimension and local extreme values \cite{b9}. The aim of this work is to use Support Vector Machine (SVM) classifier to perform the identification of weeds in relation to soybean and soil and classification of them in grass and broadleaf, aiming to apply the specific herbicide to weed detected. The critical component of any classification challenge is the available data. The publicly available dataset is used to train and evaluate machine learning algorithm. The image database collected by Dos Santos Ferreira and al. \cite{b5} contains over fifteen thousand images of the soil, soybean, broadleaf and grass weeds. As a classical pattern recognition problem, crop/weed classification primarily consists of two critical subproblems: feature extraction and classifier designation. Feature extraction is crucial step to find the suitable descriptors that can provide good discrimination between different classes. Principally, there are three main approaches for weed detection: based on color, shape and texture analysis. Various texture analysis techniques exist, texture features derived from gray-level co-occurrence matrix (GLCM) \cite{b10} and Local Binary Pattern (LBP) \cite{b11} are the most popular because of their simplicity and adaptability. In this study, color, as a primary input feature, was combined with the texture features for training SVM classifier in order to discriminate between crop/weed plants and soil.

The remainder of the paper is organized as follows. Section II describes the methodology to develop the weed detection and classification system. In section III we present the experimental results to classify weeds in soybean crop images where performance comparison between different features is carried out, and we conclude our work in section IV.

\section{METHODOLOGY}

\subsection{Image dataset}

The database used in this study is available online and can be downloaded from \textit{https://www.kaggle.com/fpeccia/weed-detection-in-soybean-crops}. It was built by Dos Santos Ferreira and al. \cite{b5}, using 400 images of soybean crop captured by the UAV. The Simple Linear Iterative Clustering (SLIC) algorithm \cite{b12} was used to segment the UAV images. The segments of each image, that identified one of the four classes used in this experiment, were annotated manually. The image dataset contained 15,336 segments, being 3249 of soil, 7376 of soybean, 3520 grass and 1191 of broadleaf weeds.

\subsection{Feature extraction}
Feature extraction is one of the most important stages in pattern recognition. It generates an input vector called descriptor for each image which is then used as input to multiclass classifiers. Although color attributes make sense in distinguishing between vegetation and Soil, they become less effective when applied to classify plant species. Sometimes, the color of weeds and crop leaves look almost the same. In this study, color, as a primary input feature, was combined with the texture features (GLCM, LBP) for discriminating soybean/weed plants and soil.

1) Color features:  The color features are means and standard deviations of the three RGB and HSV image bands.

2) Gray-Level Cooccurrence Matrix (GLCM):
Textural analysis is a very useful tool for discrimination of weeds from the main crop \cite{b13}. One of the earliest methods used for texture feature extraction was proposed by Haralick et al. \cite{b10}, known as Gray-Level Cooccurrence Matrix (GLCM) and since then it has been widely used in many texture analysis applications. 

GLCM is a second-order statistical texture analysis method. It examines the spatial relationship among pixels and defines how frequently a combination of pixels are present in an image in a given direction $\theta$ and distance \textit{d}.Various research studies show d values ranging from 1, 2 to 10. Applying large displacement value to a fine texture would yield a GLCM that does not capture detailed textural information. GLCM   directions $\theta$ of analysis are: Horizontal (0\textdegree or 180\textdegree),Vertical (90\textdegree or 270\textdegree), Right Diagonal (45\textdegree or 225 \textdegree) and Left diagonal (135\textdegree or 315\textdegree) (See Fig 1).  

\begin{figure}[htbp]
\centerline{\includegraphics[width=7cm]{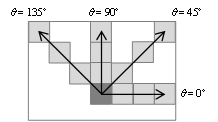}}
\caption{The direction angles for GLCM .}
\label{fig:1}
\end{figure}

Let \textit{I} be a given grey scale image. Let $ \textit{N}_{g} $ be the total number of grey levels in the image. The Grey Level Co-occurrence Matrix defined by Haralick \cite{b14} is a square matrix \textit{p}, where the $ (\textit{i},\textit{j})^{th}$ entry of \textit{p} represents the number of occasions a pixel with intensity \textit{i} is adjacent to a pixel with intensity \textit{j}. The normalized co-occurrence matrix $ \textit{p}_{d} $ is obtained by dividing each element of \textit{p} by the total number of co-occurrence pairs in \textit{p}.

The fourteen textural features proposed by Haralick et al \cite{b10} contain information about image texture characteristics such as homogeneity, gray-tone linear dependencies, contrast, number and nature of boundaries present and the complexity of the image. We used nine textural features in our study. The following equations define these features \cite{b14}.

Energy: This statistic is also called Uniformity or Angular second moment. It measures the textural uniformity that is pixel pair repetitions. Energy reaches a maximum value equal to one for a constant image.

\begin{equation}
Energy=\sum_{i}\sum_{j} \textit{p}_\textit{d}(\textit{i},\textit{j})^{2}\label{eq:1}
\end{equation}

Contrast is a measure of intensity or gray level variations between a pixel and its neighbor over the whole image. Large contrast reflects large intensity differences in GLCM. Contrast is 0 for a constant image.

\begin{equation}
Contrast=\sum_{i}\sum_{j} (\textit{i}-\textit{j})^{2}\textit{p}_\textit{d}(\textit{i},\textit{j})\label{eq:2}
\end{equation}

Entropy: This feature measures the disorder or complexity of an image. Complex textures tend to have high entropy.

\begin{equation}
Entropy=\sum_{i}\sum_{j} \textit{p}_\textit{d}(\textit{i},\textit{j})\log(\textit{p}_{d}(\textit{i},\textit{j}))\label{eq:3}
\end{equation}

Homogeneity: This feature is also called as Inverse Difference Moment. It measures image homogeneity as it assumes larger values for smaller gray tone differences in pair elements. It is more sensitive to the presence of near diagonal elements in the GLCM. Homogeneity is 1 for a diagonal GLCM.

\begin{equation}
Homogeneity=\sum_{i}\sum_{j} \frac{\textit{p}_\textit{d}(\textit{i},\textit{j})}{1+(\textit{i}-\textit{j})^{2}}\label{eq:4}
\end{equation}

Correlation: The correlation feature is a measure of gray tone linear dependencies in the image. Correlation is 1 or -1 for a perfectly positively or negatively correlated image.

\begin{equation}
Correlation=\sum_{i}\sum_{j} \textit{p}_\textit{d}(\textit{i},\textit{j})\frac{(\textit{i}-\mu_\textit{x})(\textit{j}-\mu_\textit{y})}{\sigma_\textit{x}^2\sigma_\textit{y}^2}\label{eq:5}
\end{equation}

where  $\mu_\textit{x},  \mu_\textit{y} and  \sigma_\textit{x},  \sigma_\textit{y}$ are the means and standard deviations and are expressed as:

\begin{equation}\label{eq:6}
\begin{split}
\mu_\textit{x}=\sum_{i}\sum_{j}\textit{i} \textit{p}_\textit{d}(\textit{i},\textit{j})\\ \mu_\textit{y}=\sum_{i}\sum_{j}\textit{j} \textit{p}_\textit{d}(\textit{i},\textit{j})\\
\sigma_\textit{x}=\sqrt{\sum_{i}\sum_{j}(\textit{i}-\mu_\textit{x})^2\textit{p}_\textit{d}(\textit{i},\textit{j})}\\
\sigma_\textit{y}=\sqrt{\sum_{i}\sum_{j}(\textit{j}-\mu_\textit{y})^2\textit{p}_\textit{d}(\textit{i},\textit{j})}
\end{split}
\end{equation}

The moments are the statistical expectation of certain power functions of a random variable and are characterized as follows.
Moment 1 is the mean which is the average of pixel values in an image and it is represented as

\begin{equation}
Mean=\sum_{i}\sum_{j} (\textit{i}-\textit{j})\textit{p}_\textit{d}(\textit{i},\textit{j})\label{eq:7}
\end{equation}

Moment 2 is the standard deviation that can be denoted as

\begin{equation}
Standard deviation=\sum_{i}\sum_{j} (\textit{i}-\textit{j})^{2}\textit{p}_\textit{d}(\textit{i},\textit{j})\label{eq:8}
\end{equation}

Moment 3 measures the degree of asymmetry in the distribution and it is defined as skewness

\begin{equation}
Skewness=\sum_{i}\sum_{j} (\textit{i}-\textit{j})^{3}\textit{p}_\textit{d}(\textit{i},\textit{j})\label{eq:9}
\end{equation}

Moment 4 measures the relative peak or flatness of a distribution and is also known as kurtosis:

\begin{equation}
Kurtosis=\sum_{i}\sum_{j} (\textit{i}-\textit{j})^{4}\textit{p}_{d}(\textit{i},\textit{j})\label{eq:10}
\end{equation}

3) Local binary patterns (LBP):
               
Local Binary Patterns (LBP) is a kind of gray-scale texture operator that is used for describing the spatial structure of an image texture \cite{b11}. Due to its discriminative power and computational simplicity, LBP texture extractor has become a popular approach in various applications \cite{b15}.

The original LBP operator \cite{b11} forms labels for the image pixels by thresholding the 3 × 3 neighborhood of each pixel with the center value and considering the result as a binary number. The histogram of these $2^8 = 256$ different labels can then be used as a texture descriptor.

The LBP operator was extended to use neighborhoods of different sizes \cite{b16}. Using a circular neighborhood and bilinearly interpolating values at non-integer pixel coordinates allow any radius and number of pixels in the neighborhood. Fig. 2 illustrates three neighbor-sets, where the notation (\textit{M}, \textit{R}) denotes a neighborhood of \textit{M} sampling points on a circle of radius of \textit{R}.

Given a pixel at $(\textit{x}_\textit{c}, \textit{y}_\textit{c})$, the resulting LBP can be expressed in decimal form as:

\begin{equation}
\textit{LBP}_\textit{M,R}(\textit{x}_\textit{c}, \textit{y}_\textit{c})=\sum_{i=0}^{M-1}\textit{s}(\textit{i}_{m}-\textit{i}_{c})2^{m}\label{eq:11}
\end{equation}

where $\textit{i}_{m},\textit{i}_{c}$ are respectively gray-level values of the central pixel and \textit{M} surrounding pixels in the circle neighborhood with a radius \textit{R}, and function \textit{s}(\textit{x}) is defined as:

\begin{equation}\label{eq:12}
s(x)= \left\{ 
\begin{array}{l l}
  1 & \quad \text{if $x \geq 0$}\\
  0 & \quad \text{if $x < 0$}\\ \end{array} \right.   
\end{equation}

After the LBP extraction, each pixel in an image is replaced by a binary pattern, except at the borders of the image where all of the neighbor values do not exist. The feature vector of an image consists of a histogram of the pixel LBPs. The length of the histogram is $2^{M}$ since each possible LBP is assigned a separate bin. 
In order to remove rotation effect, a rotation-invariant LBP is proposed in [16] 

\begin{equation}
LBP_{M,R}^{ri}=\min\{ROR(LBP_{M,R},i),i=0,1,...,M-1\}\label{eq:13}
\end{equation}

where $ROR(x, i )$ performs an $i$ -step circular bit-wise right shift on $x$.

Uniform Local Binary Patterns are patterns with at most two circular 0-1 and 1-0 transitions. For example, 00000000 (0 transitions) and 01110000 (2 transitions) are both uniform whereas 11001001 (4 transitions) and 01010011 (6 transitions) are not. Selecting only uniform patterns contributes to both reducing the feature dimensionality and improving the performance of classifiers using the LBP features.

\begin{figure}[htbp]
\centerline{\includegraphics[width=8cm]{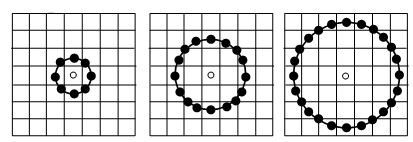}}
\caption{LBP Neighboring Pixels System.}
\label{fig:2}
\end{figure}

\subsection{Support vector machine classifier}
Once a feature descriptor is calculated, the next step deals with crop/weeds classification. Support vector machine (SVM) classifier is one of the most successful machine learning methods, because it is robust, accurate and is effective even when using a small training sample.

Support vector machines are originally developed for binary classification. But, they can be adopted to handle the multiclass classification tasks. The basic theory of SVM consists to draw an optimal hyperplane separating data points of different classes. Both separable and non-separable problems are handled by SVM in the linear and nonlinear cases. The idea behind SVM is to map the original data points from the input space to a high dimensional one, called feature space. The mapping is done by a suitable choice of Kernel function \cite{b9}.

To implement SVM on image classification we are given a certain number n of training data, each data has two parts: the d-dimensional vector of image features and the corresponding labels of classes (either +1 or -1) \cite{b17}:
\begin{equation}
E=\{(x_{i},y_{i})/x_{i}\in \mathbb{R}^d,y_{i}\in \{-1,1\}, i=1,...n\}\label{eq:14}
\end{equation}

SVM maps the d-dimensional input vector $x$ from the input space to the $d_{h}$-dimensional feature space using a nonlinear function $\varphi(.):\mathbb{R}^d \longrightarrow \mathbb{R}^{d_{h}}$. The separating hyperplane in the feature space is then defined as

\begin{equation}
w.\varphi(x)+b=0  \     /\    w \in \mathbb{R}^{d_{h}},b \in \mathbb{R} \label{eq:15}
\end{equation}

The classifier should satisfy the condition of existence of $w$ and $b$ such that:

\begin{equation}
y_{i}(w.\varphi(x_{i})+b)\geq 1\label{eq:16}
\end{equation}

However, in practical applications, data of both classes are overlapping, which makes a perfect linear separation impossible. Therefore, a restricted number of misclassifications should be tolerated around the margin. The resulting optimization problem for SVM where the violation of the constraints is penalized is given as:
\begin{equation}\label{eq:17}
\left\{ 
\begin{array}{l l l}
  \min \frac{1}{2}\|w\|^2+C \sum_{i=1}^{n}\xi_{i}  & \quad \text{such that}  \\
  y_{i}(w.\varphi(x_{i})+b)\geq 1-\xi_{i} \\
    \xi_{i}\geq 0 \\ \end{array} \right.   
\end{equation}

where $\xi_{i}$ is the relaxation factor considering classification error and $C$ is the cost parameter that controls the tradeoff between allowing training errors and forcing strict margins (i.e. empirical risk minimization \cite{b9}. 

Typically, the constrained optimization problem is referred as the primal optimization problem, which can be written in the dual space by Lagrange multipliers $\alpha_{i} \geq 0$. The solution should maximize the following expression:

\begin{equation}
\begin{split}
L(w,b,\xi_{i},\alpha_{i}) = \frac{1}{2}\|w\|^2+C \sum_{i=1}^{n}\xi_{i} \\ 
-\sum_{i=0}^{n}\alpha_{i}(y_{i}(w.\varphi(x_{i})+b)-1) \label{eq:18}
\end{split}
\end{equation}

and the dual problem is given as

\begin{equation}
\begin{array}{l l}
\max L(\alpha) =\sum_{i=1}^{n}\alpha_{i}-\frac{1}{2}\sum_{i,j=1}^{n}\alpha_{i}\alpha_{j}y_{i}y{j}K(x_{i},x_{j})\\
\text{subject to}\: 0\leq \alpha_{i} \leq C \text{and} \sum\alpha_{i}y_{i}=0 \\ \end{array}  \label{eq:19}
\end{equation}

where the Kernel function $K(x_{i},x_{j})$ corresponds to the inner product belonging to transformation space:

\begin{equation}
K(x_{i},x_{j})=\varphi(x_{i})\varphi(x_{j})\label{eq:20}
\end{equation}

There are three kind of commonly used Kernel functions: linear kernel function, polynomial kernel function and radial basis function (RBF) \cite{b18}.
Finally, the SVM classifier function can be written as

\begin{equation}
f(x)=sign(\sum_{i=1}^{n}\alpha_{i}y_{i}K(x_{i},x_{j})+b)\label{eq:21}
\end{equation}

Numerous approaches have been suggested to use SVM for multiclass classification, but there are two most popular approaches, one versus all (1.vs. all) and one versus one (1.vs.1) \cite{b19}.

The one versus all approach consists of constructing one SVM per class, which is trained to distinguish the samples of one class from the samples of all remaining classes. Usually, classification of an unknown pattern is done according to the maximum output among all SVMs. This multiclass method has an advantage that the number of binary classifiers to construct equals number of classes. However, there are some drawbacks: first, for large training data sets and increasing class number, the memory requirement is very high; second, when the number of training samples in each class is equal, the training samples sizes will be unbalanced.

The one versus one approach consists in constructing one SVM for each pair of classes. Thus, for a problem with Q classes, Q(Q-1)/2 SVMs are trained to distinguish the samples of one class from the samples of another class. Usually, classification of an unknown pattern is done according to the maximum voting, where each SVM votes for one class. The number of training data vectors required for each classifier is reduced than the previous method. Hence, this method is considered more symmetric. Moreover, the memory required is much smaller. However the main drawback of this method is the increase in the number of classifiers as the number of class increases.

\begin{figure}[htbp]
\centerline{\includegraphics[width=8cm]{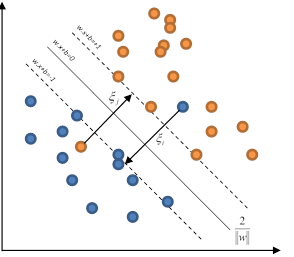}}
\caption{Illustration of SVM.}
\label{fig:3}
\end{figure}

\section{Experimental results}
To evaluate the performance of the classification system, we conducted experiments on a publicly available dataset provided by Dos Santos Ferreira and al. [5]. Totally 400 images covering four classes (100 images for each class) were extracted from the database. They include Soil, Soybean (crop), Broadleaf and Grass (weeds). A sample of the segmented images of each class used in the experiments is shown in Fig. 4.

As the primary input feature in the SVM classifier, we used the mean and standard deviation attributes of each channel of RGB and HSV color spaces.  Then the color features were combined with the texture features (GLCM and LBP) for discriminating soybean/weed plants and soil. The horizontal direction 0° with a range of 1 (nearest neighbor) was used to calculate the GLCM. Then, the nine feature values mentioned in section II (contrast, correlation, energy, entropy, homogeneity, mean, standard deviation, skewness and kurtosis) were calculated as the texture features of each image. The LBP method uses the uniform-rotation invariant LBP operator to extract texture features. The number of the neighboring pixel M is set at 8 and the radius R is set at 1.  The LBP algorithm generates a feature matrix with 10 image features. The obtained feature matrices and label values were divided randomly into two sets: the training set, and testing set, in order to train the SVM classifiers (one.vs.all and one.vs.one). The SVM kernel type is set to be Linear Basis Function. The training stage was carried out using sequential minimal optimization (SMO) algorithm \cite{b20}.

All the algorithms were developed in Matlab environment and tested on Intel Core i3 computer with a 2.53 GHz processor and 3 Gigabytes of RAM. The performance was evaluated by means of the classification accuracy, which is refer to the ability of the algorithm to predict the correct class label for instances of unknown class labels (testing set), and the confusion matrices which present the correctness of each class and the percentage of confusion of a class with the others. Because the trained data are generated randomly, the fore-coming presented scores are averaged on 10 iterations, for all experiments

Firstly, we conducted a comparison experiment to investigate the performances of the input features on SVM-based classifier. TABLE I displays the confusion matrices of the SVM classifier (1.vs.1) using different features (COLOR, COLOR+GLCM, COLOR+LBP), where the total data was divided in 70

\begin{figure}[htbp]
\centerline{\includegraphics[width=6cm]{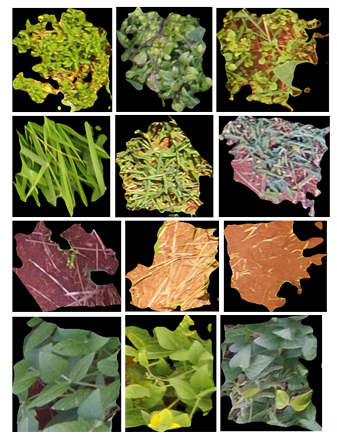}}
\caption{A sample of segmented images, from top to down: broadleaf, grass, soil and soybean.}
\label{fig:4}
\end{figure}

\begin{table}[htbp]
\caption{CONFUSION MATRICES FOR SVM (1.VS.1) BY USING DIFFERENT ATTRIBUTES}
\begin{center}
\begin{tabular}{l l l l l} 
\hline  
Classified /real &Broadleaf & Grass & Soil & Sybean \\  
\hline 
\textbf{\textit{COLOR}} \\  
 Broadleaf & \textbf{67.00} & 17.67 & 0 & 15.33\\ 
 Grass & 14.33 & \textbf{70.33} & 1.00 & 14.33\\ 
Soil & 0 & 0 & \textbf{100} & 0\\
Soybean & 7.67 & 16.33 & 0 & \textbf{76.00}\\
\textbf{\textit{COLOR+GLCM}} \\  
 Broadleaf & \textbf{72.33} & 22.00 & 0.33 & 5.33\\ 
 Grass & 15.33 & \textbf{74.33} & 0.67 & 9.67\\ 
Soil & 0 & 1.00 & \textbf{99.00} & 0\\
Soybean & 4.00 & 10.33 & 0 & \textbf{85.67}\\
\textbf{\textit{COLOR+LBP}} \\  
 Broadleaf & \textbf{95.33} & 1.00 & 0 & 3.67\\ 
 Grass & 2.33 & \textbf{95.00} & 0 & 2.67\\ 
Soil & 0 & 0 & \textbf{100} & 0\\
Soybean & 3.00 & 2.67 & 0 & \textbf{94.33}\\
\hline 
\end{tabular}
\label{tab1}
\end{center}
\end{table}

The confusion matrices given in Table I show that, the color attributes could perfectly discriminate between green vegetation (crops and weeds) and soil due to the large difference in the color information. Meanwhile, the greatest errors occurred in the classification between crops and weeds. The accuracy for the class Soil obtained by using color attributes is equal to $100$\%, i.e. is ideal. However, the classes Broadleaf, Grass and Soybean presented a large confusion. For example, $67.00$\% of data for Broadleaf class are correctly classified, $17.67$\% are mismatched with Grass class and $15.33$\% are confused with Soybean class. This confusion was slightly reduced by adding GLCM Texture information to color spaces ($72.33$\% of Broadleaf are correctly classified). The classification was improved by the combination of color attributes with LBP features ($95.33$\% of broadleaf are correctly classified). Examining the results more closely, we can say that the LBP method produces more consistent classification accuracy performance, in comparison with the GLCM features. The COLOR+LBP was the best combination of input features for the SVM models for the classification. Therefore, the COLOR+LBP features were used as the final model for further analysis.

On the other hand, a performance comparison between one versus all and one versus one approaches
using COLOR+LBP features was established. The results of the overall classification accuracies and the
prediction times for the two approaches with different numbers of input images in training and testing phases are shown in Tables II and III.

\begin{table}[htbp]
\caption{TOTAL ACCURACY FOR SVM CLASSIFIERS USING COLOR+LBP FEATURES}
\begin{center}
\begin{tabular}{|c|c|c|c|}
\hline
\textbf{Method}&\multicolumn{3}{|c|}{\textbf{Training set percentage}} \\
\cline{2-4} 
 & \textbf{\textit{30\%}}& \textbf{\textit{50 \%}}& \textbf{\textit{70 \%}} \\
\hline
SVM one.vs.all& 89.43\%& 91.05 \%& 93.92 \% \\
\hline
SVM one.vs.one& 92.89\%& 94.35 \%& 96.17 \% \\
\hline
\end{tabular}
\label{tab2}
\end{center}
\end{table}

\begin{table}[htbp]
\caption{COMPUTATIONAL TIME  FOR SVM CLASSIFIERS USING COLOR+LBP FEATURES}
\begin{center}
\begin{tabular}{|c|c|c|c|}
\hline
\textbf{Method}&\multicolumn{3}{|c|}{\textbf{Training set percentage}} \\
\cline{2-4} 
 & \textbf{\textit{30\%}}& \textbf{\textit{50 \%}}& \textbf{\textit{70 \%}} \\
\hline
SVM one.vs.all& 1.2966 s& 1.5382 s& 1.8442 s \\
\hline
SVM one.vs.one& 0.6242 s& 0.8580 s& 1.2206 s \\
\hline
\end{tabular}
\label{tab3}
\end{center}
\end{table}

From Tables II and III, We remark that the performance ratio increases with the number of of input images. The SVM classifier with one.vs.one approach is promising in terms of overall accuracy and algorithm speed where, it presents the most outstanding performance compared to the one.vs. all method. The overall classification accuracy, using one .vs. one approach reaches $96.17$\% yields over $2$\% higher accuracy than the one .vs. all approach (i.e., $93.92$\%).   As shown in execution times, the SVM one.vs.one strategy is much faster than one.vs.all approach. Indeed, the training time of SVM classifier increases significantly with the number of training samples. Thus, since the number of samples which are needed to train the SVM classifier of one .vs. one strategy become smaller, it is generally faster to train the $6$ SVMs of the one.vs.one method than the $4$ SVMs of the one .vs. all approach. 

\section{Conclusion}
In this work, we have applied SVM classifier on UAV images to discriminate crops, weeds and soil. We have evaluated the following input features: color attributes, GLCM texture features and LBP extractor.  The confusion matrices have been obtained for different combinations. We notice that the misclassifications occurred mainly between crop and weed classes, and these were improved by adding Texture information to color space. Soil could be accurately discriminated using only color features since soil has a strong color difference with green vegetation. The LBP extractor combined with color features produce more consistent classification accuracy performance. The use of SVM classifier with one versus one approach achieved excellent results, with accuracy higher than $96$\% in the classification of all classes, and is computationally effective.

In future, we can explore other features for training the classifiers and analyze the effects of other machine learning algorithms for classifying crop images. Particularly, we investigate the use of deep learning to reduce the confusion between crops and weeds.

\end{document}